\documentclass[letterpaper, 10 pt, conference]{ieeeconf}  

\IEEEoverridecommandlockouts                              

\usepackage{graphicx}
\usepackage{amsmath}
\usepackage{amssymb}
\usepackage{booktabs}
\usepackage{xcolor}

\title{\LARGE \bf
MoEmo Vision Transformer: Integrating Cross-Attention and Movement Vectors in 3D Pose Estimation for HRI Emotion Detection}

\author{David C. Jeong$^{1}, \textit{Member, IEEE}$, Tianma Shen$^{1}$, Hongji Liu$^{1}$, Raghav Kapoor$^{1}$, Casey Nguyen$^{1}$, \\Song Liu$^{2}, \textit{Member, IEEE}$, Christopher A. Kitts$^{1}$, \textit{Senior Member, IEEE}
\thanks{$^{1}$David Jeong (dcjeong@scu.edu), Tianma Shen (tshen2@scu.edu), Hongji Liu (hliu4@scu.edu), Raghav Kapoor (rkapoor@alumni.scu.edu), Casey Nguyen (cnguyen5@scu.edu), and Christopher Kitts (ckitts@scu.edu) are with Santa Clara University.
        }%
\thanks{$^{2}$Song Liu (liusong@shanghaitech.edu.cn) is with the School of Information Science and Technology, ShanghaiTech University. 
        }%
}

\begin{document}

\maketitle
\thispagestyle{empty}
\pagestyle{empty}

\begin{abstract}

Emotion detection presents challenges to intelligent human-robot interaction (HRI). Foundational deep learning techniques used in emotion detection are limited by information-constrained datasets or models that lack the necessary complexity to learn interactions between input data elements, such as the the variance of human emotions across different contexts. In the current effort, we introduce 1) MoEmo (Motion to Emotion), a cross-attention vision transformer (ViT) for human emotion detection within robotics systems based on 3D human pose estimations across various contexts, and 2) a data set that offers full-body videos of human movement and corresponding emotion labels based on human gestures and environmental contexts. Compared to existing approaches, our method effectively leverages the subtle connections between movement vectors of gestures and environmental contexts through the use of cross-attention on the extracted movement vectors of full-body human gestures/poses and feature maps of environmental contexts. We implement a cross-attention fusion model to combine movement vectors and environment contexts into a joint representation to derive emotion estimation. Leveraging our Naturalistic Motion Database, we train the MoEmo system to jointly analyze motion and context, yielding emotion detection that outperforms the current state-of-the-art.  


\end{abstract}

\section{INTRODUCTION} 

Recent years have observed an increased demand for the development of robots capable of working alongside and interacting with humans, with a high prioritization of (human) safety~\cite{ruffaldi2016third}. The development of these cobots has enabled the unprecedented sharing of work spaces between robots and human collaborators~\cite{ajoudani2018progress}. Such human-robot collaboration is contingent on modeling human communication in-the-wild. Such human-human interactions largely involve the processing of signals and cues, such as speech and nonverbal behavior. In the process of learning the social meanings of such cues, psychologists traditionally study emotions.

Affect recognition is a critical facet of machine perception that has broad impacts on human-robot interaction (HRI)~\cite{tuyen2018}, which may afford \textit{functional} solutions in human-robot collaboration~\cite{ajoudani2018progress}, as well as \textit{social} solutions in human-robot interaction~\cite{dautenhahn2007socially}, such as in care giving~\cite{bemelmans2012socially}. For instance, accurate emotion perception from a robot is associated with human comfort within HRI~\cite{park2012law}.

Towards addressing understanding of emotion, we observe prominent research work within the field of computer vision (CV) through visual emotion analysis (VEA)~\cite{yang2022seeking,xu_mdan_2022}. Much of this research investigates the role of facial expressions in emotions, but emotion detection in-the-wild under uncontrolled conditions present challenges, such as heavy reliance on image clarity and quality.

Further, prior CV works on visual emotion analysis only handles prediction based on images of subjects, thus missing valuable information gained from sequences of frames for real-time evaluation of subjects of interest to robustly model implicit signals from non-verbal human communication. Thus, such works may not be sufficient for comprehensive and accurate emotion detection irobots~\cite{bemelmans2012socially}. 

Thus, the current work employs attention mechanisms in its analysis of sequential image data and contextual emotion detection based on human body motion. Our main contributions include: (1) MoEmo (Motion to Emotion), a vision transformer model for emotion detection of 3D \textit{body} motion that is flexible across eight social situational contexts. (2) An original database of human body movement videos that afford context-dependent emotion detection. 

We structure the paper as follows: First, we provide a review of related work regarding emotion and context, pose estimation models, emotion detection models, and emotion detection databases. Then, we introduce the technical components of our proposed model. Next, we introduce our database, including data collection procedures. Finally, we compare our results relative to existing benchmarks.

\vspace{-.1pc}
\section{RELATED WORK}
\vspace{-.2pc}

\subsection{Emotion Detection}

Emotion detection research~\cite{10.1109/cvpr52688.2022.02026} typically utilizes a six-category framework of emotions including joy, anger, disgust, fear, sadness, and surprise~\cite{ekman1972hand}, as well as the Valence-Arousal-Dominance (VAD) model~\cite{kosti2017emotion} of emotion. Emotion recognition among humans often relies on nonverbal cues, but may also be impacted by context~\cite{miller2019causal}. Although emotion research typically relies on facial expression cues~\cite{ekman1972hand}, the current work focuses on physical body movements. Nonverbal behaviours for emotional expression account for 93\% of the human communication~\cite{mazhar2018} and figures prominently in human-robot interaction~\cite{nakata1998expression}.
\subsection{Context in Emotion Detection}
A critical grounded assumption of this work is that human social judgments (e.g, emotions) based on cues (e.g., body pose) are subject to change across different contexts (e.g., physical environment)~\cite{rauthmann2014situational, miller2019causal}. The DIAMONDS taxonomy~\cite{rauthmann2014situational} is the leading psychological framework to categorize social situational contexts. HRI research often analyzes behavioral cues, such as social distance (e.g., proxemics)~\cite{breazeal2009embodied,takayama2009influences,mumm2011human}. While fewer emotion detection research in HRI address context-based analysis of social cues, recent developments within computer vision account for context in emotion detection~\cite{huang2021emotion,kosti2019context}.



\vspace{-.5pc}
\subsection{Emotion Detection in Computer Vision}

\subsubsection{Pose Estimation}

Pose estimation~\cite{10.1109/iccvw54120.2021.00403} is a small but critical component to emotion detection within computer vision. Pose estimation refers to the process of localizing anatomical keypoints on human bodies in images and videos~\cite{cao2017realtime} and largely focuses on detecting human body joints~\cite{pishchulin2012articulated} for outcomes such as motion action representation~\cite{Butepage2017CVPR}, action synthesis~\cite{zhou2018auto}, and motion prediction~\cite{Kanazawa2019CVPR}. Pose estimation techniques locate these body joints after identifying humans in image(s) to extract meaningful feature maps of the body poses~\cite{10.5220/0010359506690679}.

P-STMO (Pre-Trained Spatial Temporal Many-to-One Mode)~\cite{PSTMO, shan2022p} is a robust pose estimator model that affords conversion of human body movements in 2D videos into a 3D representation of body movement, thereby yielding more accurate spatial and temporal information relative to 2D pose estimators such as AlphaPose\cite{li2021hybrik,fang2022alphapose} and OpenPose~\cite{8765346}.\footnote{AlphaPose recently accomplished 3D pose estimation~\cite{fang2022alphapose}.} In the current work, we leverage P-STMO alongside AlphaPose due to its documented higher performance over OpenPose on both the COCO dataset (18\% higher than OpenPose) and the MPII dataset (8.5\% higher than OpenPose). The STCFormer~\cite{Tang_2023_CVPR} introduces a novel transformer architecture that employs cross-attention transformer blocks to integrate spatial and temporal information, achieving superior performance over the above state of the art methods.~\footnote{This work~\cite{Tang_2023_CVPR} was published after the review stage of the current work.}

\subsubsection{Existing Emotion Detection Models in Computer Vision}
Emotion detection models in computer vision have largely focused on extracting information from facial expressions~\cite{10.1109/lra.2019.2930434}. While these works have demonstrated non-trivial levels of efficacy in emotion detection, we believe that such algorithms can improve by accounting for full-body nonverbal behavior. In addition to noting observation in normative human-human social situations, we also draw on works that use nonverbal communication such as poses and gestures~\cite{stoeva2020,10.1109/iciev.2019.8858536} as they demonstrate a significant development in emotion detection  within computer vision.

Although neural networks have achieved great advances in computer vision, the current state-of-the-art models also face two major problems. First, these models heavily rely on typical vision data sets. Typically researchers must create and label millions of data to obtain strong results, which is very labor-intensive and costly. The second problem is that standard vision models lack flexibility. 

\paragraph{Vision Transformer (ViT).}
To solve the above two problems, CLIP~\cite{radford2021learning} is a pre-trained Vision Transformer (ViT) that offers superior learning advantages relative to neural networks. Vision Transformers such as CLIP efficiently learn visual concepts from a wide variety of images and natural language supervision to derive richer representations for vision tasks, thus enabling it to be applied to any visual classification benchmark. Ultimately, CLIP affords a strong feature map based on a significantly smaller and cost-efficient database. In the following, we offer prior (albeit limited) solutions to emotion detection within computer vision to support our application of ViT.
            
\paragraph{ResNet50.} The ResNet50~\cite{resnet} architecture is often used due to its ability to extract high-level information and ensure stable training through the use of residual connections. However, recent efforts that leverage various attention mechanisms appear to outperform the vanilla ResNet50 architecture~\cite{attention2017}. For example, we observe that the WSCNet model ~\cite{Yang_2018_CVPR} improves upon the ResNet50 by adding spatial attention to the bottom layers. While this model yields improved performance, the feature maps learned by the model are missing important channel information, which limits the model's performance. 

\paragraph{PDANet.} Alternative to ResNet50 and WSCNet, the PDANet model~\cite{zhao2019pdanet} adds channel-wise attention and spatial attention to ResNet50 to improve the accuracy of the classification. However, the richness of the representations learned by such an attention model is not as comprehensive as that of the Transformer model as the latter retains a longer range that is capable of extracting greater information. Notably, less information is stored with input data made up of individual images than that of videos, which are essentially sequences of image frames.

\paragraph{ABAW.} The ABAW model~\cite{kollias2022abaw} resolves the issue noted above by not only extracting information from video input (i.e., sequential image frames) but also by considering the respective context of each of the videos. However, ABAW is limitated by the use of a the LSTM model to deal with the sequence of frames that is later concatenated with the other input data. The inherent design of the LSTM neuron leads to training instabilities because of the challenges faced by using the backpropagation algorithm, which ultimately leads to the concatenation process losing valuable information for capturing the relationship between the original input data and the extracted context.
            
\paragraph{BCEmotion.} To resolve the above issues observed with LSTM, the BCEmotion model~\cite{10.1109/iccvw54120.2021.00403} manages the video input through the use of the 3D ResNet architecture. Although this model demonstrates better performance than previous work, it does not attain accuracy comparable to that of the Vision Transformer model. Additionally, the BCEmotion model applies the feature concatenation operation (as in ABAW) to combine two different types of feature maps which leads to dips in performance.

\subsubsection{Body Motion Databases}

Towards modeling the complex task of emotion detection, we note few existing public databases where visual input of human movement is labeled with the respective emotion. In~\cite{10.5220/0010359506690679} for instance, the EMOTIC database~\cite{10.5220/0010359506690679}, a collection of images of people annotated according to respective emotional state, is introduced. However, the lack of additional information that could be retrieved from multiple sequential frames limits the capabilities of models learning from EMOTIC. 

Additionally, we recognize other databases such as the Body Language Dataset (BoLD)~\cite{luo2021towards}, MPI Emotional Body Motion Database~\cite{volkova2014mpi}, and CMU Graphics Lab Motion Capture Database~\cite{van2013biological} where annotations of short-video clips of human movement to emotion are provided. However, these databases tend to lack sequential image frames (e.g., videos) of whole-body movement alongside labels of emotions that \textit{also} account for context. Specifically, while the BoLD annotates videos of human movement to emotion, these videos are limited to upper body shots, thus constraining full-body interpretations. And while the MPI Emotional Body Motion Database and CMU Graphics Lab Motion Capture Database do include full-body annotation of emotions, these datasets are purely skeletal (not actual human video) in nature, thus lacking context from consideration. Finally, the BoLD dataset contains an excess of different contexts and body movements that lack adequate control for our purposes. 

\section{MoEmo Transformer}
\label{sec:formatting}
\begin{figure*}[!htbp]
    \centerline{\includegraphics[scale=0.13]{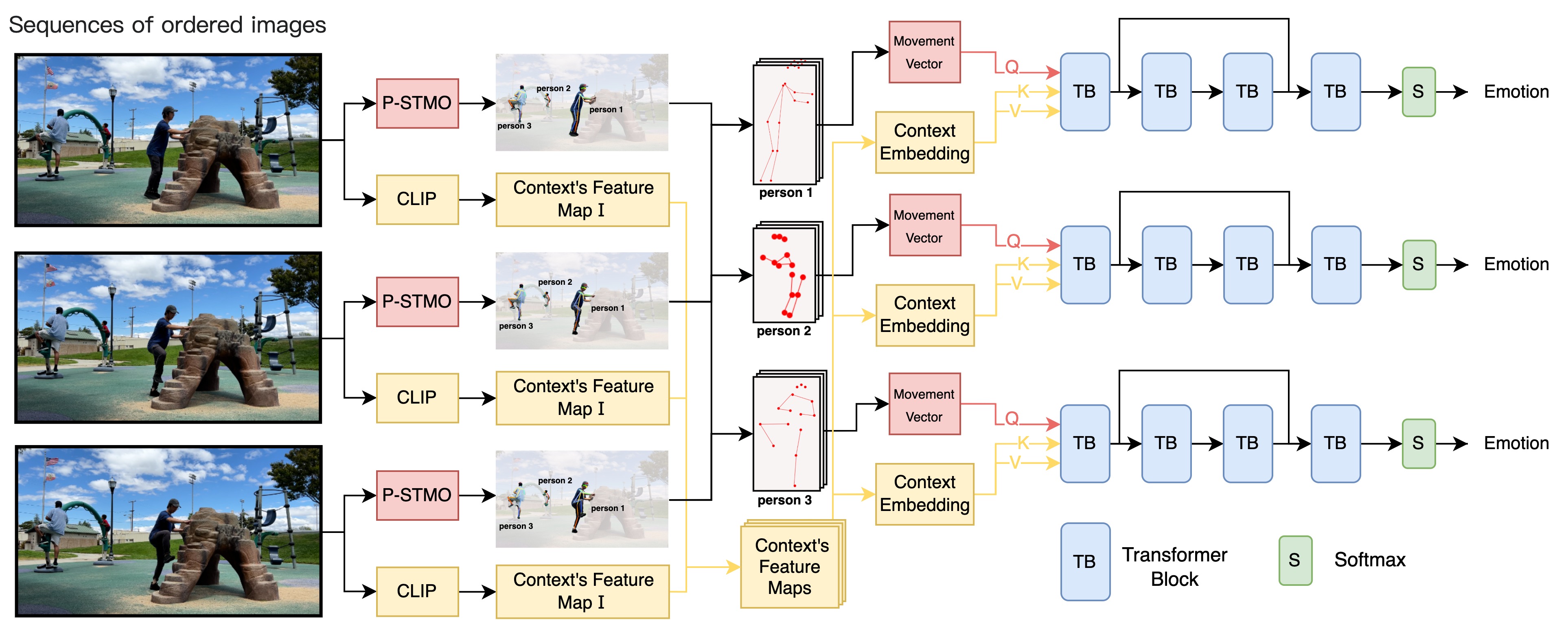}}
    \caption{Complete structure of MoEmo, where TB (Transformer Block), includes MLP (Multilayer Perceptron) and normalization layers; S (softmax function). The input video (left) is re-sampled at a frequency rate of 4Hz. Movement vectors and feature maps of contexts extracted for each frame.}
    \label{MoEmo}
\end{figure*}

In the following we introduce an effective framework for emotion detection in the wild. To estimate human emotion under general circumstances, we extract features contributing to emotions based on the body posture of individuals, which is derived by identifying key skeletal points on the subject of interest and the context. From this information, emotion detection is then performed by combining the prior features. Figure ~\ref{MoEmo} shows the overall framework of the emotion detection algorithm based on real-life video.

\subsection{Context Feature Extraction Module} 

CLIP utilizes both an image encoder and a text encoder to predict which image pairs, of which we utilize the image encoder to obtain the feature map of \textit{context}. 
The output of CLIP is the feature map of images $I_{i}$, where $i$ is the index of frames of the input video. The shape of the feature map $I_{i}$ is $b \times 50 \times 768$, where $b$ is the batch size. After the last layer of CLIP, we propose adding a context embedding block. The block consists of a concatenation layer to concatenate context's feature maps and convolutional layers to transform the shape of the context's feature map $f_{c}$ onto the feature map of the human body movement shapes, $V$, as shown in Figure ~\ref{MoEmo}. Each person shares the same context feature maps. The convolution embedding block can also generate the different contexts for each person in the frames, which is time-efficient, because the context's feature map $f_{c}$ solely requires one computation, regardless of the number of people.

\subsection{Movement Vector Estimation}

To capture the feature map of body movements, we compute body movement vectors of human 3D skeleton joints.
We produce the movements vectors $V$ by first applying the P-STMO (Pre-Trained Spatial Temporal Many-to-One Mode) model ~\cite{PSTMO} to gain 17 keypoints of 3D skeleton joints, which includes ears $L^{1},L^{5}$, eyes $L^{2},L^{3}$, nose $L^{4}$, shoulders $L^{8},L^{9}$, elbows $L^{7},L^{10}$, hands $L^{6},L^{11}$, wrists $L^{14},L^{15}$, knees $L^{13},L^{16}$ and ankles $L^{12},L^{17}$. To solve the multi-person task, we first apply the YOLO Model (You Only Look Once)~\cite{redmon2016you} to detect the humans in the image frames within a bounding box. Next, P-STMO computes each person in the frames as input data, one by one. The shape of P-STMO's output is $p \times f \times 17 \times 3$, where $p$ is the number of people in the frames; $f$ is the number of frames; three represents three dimensions of $x,y,z$. We split the output for each person to detect people's emotions respectively. 


Based on 3D joints, the second step is to determine pairs of pixels across image frames for the movement vectors $V$. We can seamlessly address this step due to the collection of 3D joint keypoints, which provide a clear reference point of not only each body joint, but also each vector across each image frame. 3D joint keypoints across multiple image frames can be represented as $L_{i}=(x_{i},y_{i},z_{i})$, where $i$ is the index of frames, and each pair of 3D joints (e.g., elbow in frame 1, 2...) across multiple image frames is $L_{i}, L_{i+1}$. 
\begin{equation}
    \begin{aligned}
    V_i &=\left(L_i, L_{i+1}\right) \\
    &=\left(\left(x_i, y_i, z_i\right),\left(x_{i+1}, y_{i+1}, z_{i+1}\right)\right)
    \end{aligned}
\end{equation}
As such, we present a movement vector $V_{i}$ shown in Equation 1, where frame 1 (e.g., initial frame) is represented by the point $(x_{i},y_{i},z_{i})$ and subsequent frames are represented by the point $(x_{i+1},y_{i+1},z_{i+1})$. Thus, the shape of each person's movement vector is $(f-1) \times 17$ (number of body joint keypoints) $ \times 6$, reflecting the three dimensions of x, y, z across each pair of frames analyzed in the movement vectors. 

\subsection{Cross-attention Transformer}

We integrated a Cross-Attention Transformer shown in Figure ~\ref{MoEmo}, which asymmetrically combines two separate embedding sequences of the same dimension (context feature maps, movement vectors). In contrast, the self-attention input is a single embedding sequence. The movement vectors $V$ serve as a query input, while the context's feature map $C$ serve as key and value inputs. Emotion detection is represented by the emotion probability model $P_{\theta}(V_{0:T}|C_{0:T})$ where $V_{0:T}$ represent $V_{1}, V_{2}, ... , V_{T}$ movement vectors of T frames re-sampled from the input video. $C_{0:T}$ represent $C_{1}, C_{2}, ... , C_{T}$ contexts of T frames; $\theta$ represents the parameters to be trained in the MoEmo Transformer. That is: 
\begin{equation}
    \begin{aligned}
    P_{\theta}(V_{0:T}|C_{0:T})&=\prod ^{T}_{i=0}P_{\theta }\left( V_{i}\vline C_{0:T}\right)\\
    &=\prod ^{T}_{i=0}P_{\theta }\left( C_{0:T}\vline V_{i}\right) \times \dfrac {P_{\theta }\left( V_{i}\right) }{P_{\theta }\left( C_{0:T}\right) }\\
    &=softmax\left( \dfrac {q_{V}\cdot k^{T}_{C}}{\sqrt {n}}\right) .v_{C}
    \end{aligned}
\end{equation}
In Equation 2, $q_{V}$ represents the query of the cross-attention transformer, which represents the condition of the probability $P_{\theta }\left( C_{0:T}\vline V_{i}\right)$. $k^{T}_{C}$ is the key of the cross-attention transformer, which is the latent variable. $v_{C}$ is the value of the cross-attention transformer, which is the weight $\prod ^{T}_{i=0} \dfrac {P_{\theta }\left( V_{i}\right) }{P_{\theta }\left( C_{0:T}\right) }$ of the emotion probability. $n$ is the size of $q$.

As observed in Equation 2, two different inputs can extract information from the local neighborhoods by attention weights $A = softmax\left( \dfrac {q_{V}\cdot k^{T}_{C}}{\sqrt {n}}\right)$ based on the pairwise similarity between two elements of the sequence, namely their respective query $q_{V}$ and key $k^{T}_{C}$ representations. The dot product can pay attention to the connection between body movements and context. Thus, the MoEmo Transformer can highlight salience in each 3D motion of joints corresponding to the surrounding context, ultimately impacting emotion.

Based on the success of the residual transformer blocks~\cite{liu2021swin}, we apply the residual blocks that share the same weights (across each of the layers) in the middle of the transformer blocks to increase the robustness of the network, as shown in Figure ~\ref{MoEmo}. At the final layer (i.e., after the output of all transformer blocks), we apply the softmax function to obtain the emotion probability distribution.
\section{Naturalistic Motion Database}

To address the limitations of the aforementioned datasets (e.g., EMOTIC~\cite{10.5220/0010359506690679}, BoLD~\cite{luo2021towards}, MPI Emotional Body Motion Database~\cite{volkova2014mpi}, and CMU Graphics Lab Motion Capture Database~\cite{van2013biological}), we introduce the \textit{Naturalistic Motion Database} to account for full-body human movement videos that include emotion labels and controlled naturalistic social contexts. We achieve this control in video context based on a) videos occurring in their natural environment and b) videos with a green screen studio backdrop that affords chroma keying for both a wide variety of environments (See Fig.~\ref{example}).



\subsection{Procedure}

Data collection involved filming individual clips of 15 diverse subject actors in a professional film and TV production studio.\footnote{Our dataset only includes single person shots in single frames, which is also why we are unable to test for multi-person detection).} To ensure the quality of the recorded videos, we used the following film studio equipment: Two Lowel Fresnel 650 Key Lights, Two Strand Lighting-Quartz Color IRIS 1 bulbs, Four Strand Lighting-Quartz Color IRIS 2 bulbs, Sony Alpha 3 and Digital Waterproof Green Screen paint. Each subject performed poses at a specific speed commensurate with the intended emotion, and video recordings of these poses were taken from three different angles. 
We identified eight background types (as context) to chroma key over the green screen corresponding to the DIAMONDS taxonomy~\cite{rauthmann2014situational}, namely work-related, learning-based/thinking, threatening, relationship/dating, positive, negative, deception, and social. 

\begin{figure}[htbp]
    \centerline{\includegraphics[scale=0.11]{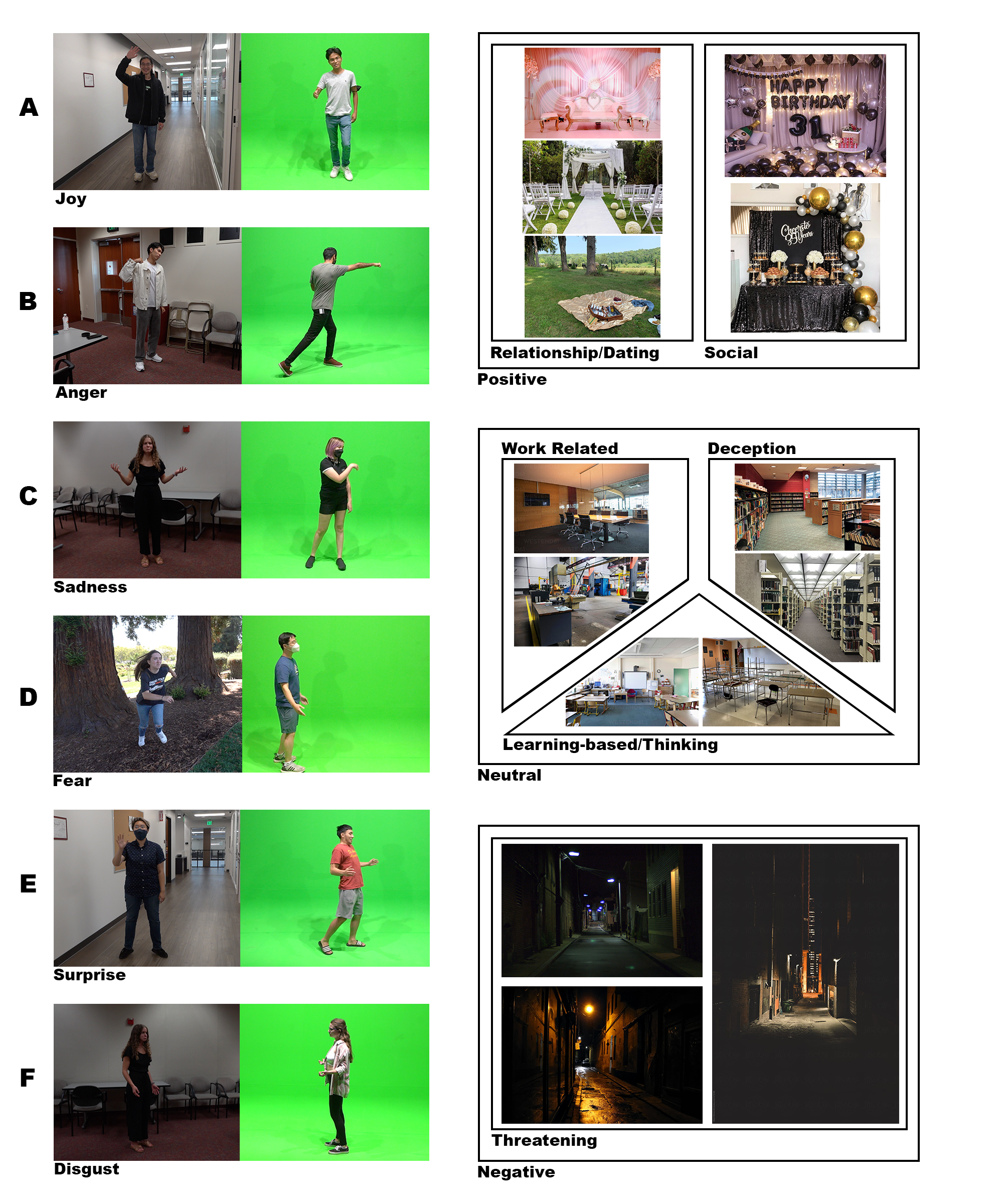}}
    \caption{Left includes emotion samples of the Naturalistic Motion Database. Right includes contexts for the green screen videos. } 
    \label{example}
\end{figure}

Based on these background/context categories, we used a web crawler to obtain a random selection of diverse, high-resolution (at least 480P) images of our background types, including birthday (39), proposal (52), beach (18), picnic (45), restaurant (37), wedding (100), classroom (21), industry (74), library (45), dark alley (65), under the bridge (65), office (19), meeting room (74), lecture hall (37). Once obtained, images were subjected to cleaning to ensure the successful integration of the backgrounds with human subjects. First, we filtered out images that were crowded with too many individuals. We also removed any background images that would introduce occlusions over the subject of interest. Subsequently, we omitted any backgrounds containing watermarks that would negatively impact the quality of our video. Based on the labels of green screen videos with the different background images, we collapsed the eight context categories into three composite categories, namely Negative, Neutral, and Positive, as seen in Fig ~\ref{example}.


The image search results (emotions) were then overlaid with all of backgrounds. All emotion types were overlaid with all context types, and we ensured to generate a balanced distribution between each emotion category. Currently, the dataset consists of 1512 high-resolution videos, as seen in Table~\ref{tab:my-table}. From this dataset, we selected 10\% and 90\% for the testing and training sets, respectively.
\begin{table}[]
    \caption{Naturalistic Motion Database}
    \label{tab:my-table}
    \resizebox{\columnwidth}{!}{%
    \begin{tabular}{@{}lccccccc@{}}
    \toprule
    Dataset  & Angry & Disgust & Fear & Joy & Sadness & Surprise & ALL  \\ \midrule
    Whole    & 255   & 86      & 334  & 219 & 358     & 260      & 1512 \\
    Training & 229   & 77      & 301  & 197 & 322     & 233      & 1359 \\
    Testing  & 26    & 9       & 33   & 22  & 36      & 27       & 153  \\ \bottomrule
    \end{tabular}%
    }
\end{table}

\section{Experiments and results}

\subsection{Evaluation Metrics}
To evaluate the visual emotion regression performance, we report both accuracy as well as F1 scores. We compute accuracy both in terms of individual six emotion categories, as well as the composite means of all categories. As seen in Table ~\ref{tab:my-table}, the image sample size for disgust is disproportionately smaller than the remaining emotion categories, thus warranting the computation of an F1 score to estimate the comprehensive performance of all emotion categories.

\begin{table*}[]
    \caption{Performance comparison measured by accuracy and F1 score on the testing  Naturalistic Motion Database}
    \label{tab:2}
    \resizebox{\textwidth}{!}{%
    \begin{tabular}{@{}ccccccccc@{}}
        \toprule
        Model                     & Accuracy       & F1 score       & joy            & angry          & disgust        & fear           & sadness        & surprise       \\ \midrule
        ResNet50                  & 0.784          & 0.674          & 0.909          & 0.769          & 0.667          & 0.727          & 0.833          & 0.741          \\
        Vision Transformer\cite{dosovitskiy2020image}        & 0.849          & 0.809          & 0.864          & 0.808          & 0.778          & 0.818          & 0.861          & 0.889          \\ \midrule
        WSCNet\cite{Yang_2018_CVPR}                    & 0.804          & 0.687          & \textbf{0.955} & 0.846          & 0.667          & 0.727          & 0.833          & 0.741          \\
        PDANet\cite{zhao2019pdanet}                    & 0.824          & 0.699          & \textbf{0.955} & 0.885          & 0.667          & 0.758          & 0.861          & 0.741          \\
        ABAW\cite{kollias2022abaw}                      & 0.837          & 0.724          & 0.864          & 0.885          & 0.667          & 0.788          & 0.917          & 0.778          \\ \midrule
        MoEmo w/o context         & 0.758          & 0.671          & 0.864          & 0.769          & 0.556          & 0.697          & 0.806          & 0.741          \\
        MoEmo w/o cross-attention & 0.863          & 0.831          & 0.864          & 0.885          & 0.778          & 0.879          & 0.861          & 0.852          \\
        MoEmo                     & \textbf{0.915} & \textbf{0.887} & \textbf{0.955} & \textbf{0.923} & \textbf{0.889} & \textbf{0.909} & \textbf{0.889} & \textbf{0.926} \\ \bottomrule
    \end{tabular}%
    }
\end{table*}

\subsection{Comparison with the State-of-the-art}
We trained and tested our MoEmo Model and three state-of-the-art approaches on the same training and testing datasets, respectively, as shown in Table ~\ref{tab:my-table}. Two of the current state-of-the-art models (WSCNet, PDANet) utilize images as input data over video. Thus, for training purposes, we selected three of the 16 image frames from our video data as inputs. For testing purposes, we select one image frame that yields the highest performance among the three for testing. More specifically, we selected the first (1st), middle (8th), last (16th) out of the 16 frames, which we then input to models for comparisons.

The MoEmo Model achieved the highest performance (F1 scores and accuracy) relative to three state-of-the-art models (WSCNet, PDANet, ABAW) and two seminal models (ResNet50, Vision Transformer) across all emotion categories, as seen in Table~\ref{tab:2}. The first section in the Table~\ref{tab:2} reports the results of the seminal models. We pre-trained the Vision Transformer using CLIP to extract the input frames into feature maps. CLIP retains good features of a transformer and is trained on a robust dataset that is approximately 1000 times larger than ImageNet, thus yielding a feature map that is higher in quality than that of PDANet. 

The second section in Table~\ref{tab:2} reports the results of the state-of-the-art models. WSCNet is based on the ResNet50 model, with the addition of spatial attention at the latter part of the model. PDANet improves upon WSCNet by adding channel-wise attention to enlarge the view of the model. Naturally, both WSCNet and PDANet's accuracy and F1 scores are higher than ResNet50. Although PDANet contains channel-wise and spatial attention, CLIP's transformer blocks are more efficient than channel-wise and spatial attention because of the presence of the multi-layer perceptron and the weights of the query, key, and value. In addition to the limitation of the attention modules (channel-wise, spatial attention), PDANet is limited to processing images as input instead of videos. Finally, ABAW does process videos as inputs (utilizing LSTM) as well as consider the environmental context in the networks, but for the multi-input sequence, ABAW solely utilizes the concatenate layer to connect the different feature maps, which leads to temporal information loss. The differences observed between ABAW and MoEmo demonstrate that the cross-attention utilized in MoEmo is more robust in modeling temporal relationships, such as the one between human pose and social context. 

\subsection{Ablation Study}
The proposed MoEmo Transformer Model contains two novel contributions: 1) Accounting for context information within emotion detection and 2) Leveraging a cross-attention mechanism across two different feature maps. In Table~\ref{tab:2}, variations on the MoEmo model are shown, in which one does not include context information, and analysis is performed only on the 17 keypoints of human skeletons to compute human emotion. The performance deficiency observed in the MoEmo model without context may be attributed to the loss of valuable contextual information from the input data, constraining emotion detection accuracy.

Furthermore, we present the results of the MoEmo transformer with the cross-attention removed, utilizing a concatenation layer instead to combine the movement vectors and context feature maps. The result of this variation of MoEmo without cross-attention demonstrates that cross-attention can extract more information than the concatenation layer. In addition, when compared with Vision Transformer, the MoEmo without cross-attention model demonstrates observable, non-trivial benefits from analyzing multi-length sequence data (e.g., video frames) compared to a single image. In the case of removed cross attention, the model needs a concatenation layer before the conventional transformer steps. Removing context will not remove the \textit{kv} value because it is essential for an attention mechanism. Thus, when we remove cross attention, it reverts to vanilla soft attention.

\section{CONCLUSION}
In this work, we demonstrate the superior performance of the MoEmo system compared to the state-of-the-art works in emotion detection. Upon closer investigation into the design of MoEmo, we attribute performance improvements to the use of cross-attention, feature maps from the CLIP encoder for extracting context information and optimal flow data from the derived 3D skeletal information. Further, we note the contribution of the Naturalistic Motion Database as a comprehensive public database for the emotion detection task. In addition to recording a diverse array of subjects of interest, our data collection methodology provides a flexible and straightforward means of accounting for a greater variety of in-the-wild contexts through synthetic means. Additionally, further advancement in the use of cross-attention demonstrates promise given the current performance, in which further improvements in the attention mechanism may capture more meaningful relationships between the extracted optimal flow information and context embedding vector. A pressing limitation of this work lies in the size of our dataset. Due to the complexity of the emotion detection task, we believe that adding a greater variety of contexts and subjects of interest will result in future models learning more meaningful subtleties in nonverbal communication vital for the task at hand. Due to the limited size of the dataset, we cannot conclude that the MoEmo Transformer System accurately detects emotion from all types of body movement, but the current results demonstrate that our model supports flexible emotion detection for functional human-robot collaboration. Further limitations may include a potential inadequacy of this model for real-time human-robot interaction. In our future work, we anticipate adding to our database in specific HRI contexts (e.g., care-taking, medicine) to further improve the performance of MoEmo while maintaining human-centered design and social responsibility.

\bibliographystyle{plain}
\bibliography{references}

\begin{thebibliography}{10}

\bibitem{ajoudani2018progress}
Arash Ajoudani, Andrea~Maria Zanchettin, Serena Ivaldi, Alin Albu-Sch{\"a}ffer,
  Kazuhiro Kosuge, and Oussama Khatib.
\newblock Progress and prospects of the human--robot collaboration.
\newblock {\em Autonomous Robots}, 42(5):957--975, 2018.

\bibitem{bemelmans2012socially}
Roger Bemelmans, Gert~Jan Gelderblom, Pieter Jonker, and Luc De~Witte.
\newblock Socially assistive robots in elderly care: a systematic review into
  effects and effectiveness.
\newblock {\em Journal of the American Medical Directors Association},
  13(2):114--120, 2012.

\bibitem{breazeal2009embodied}
Cynthia Breazeal, Jesse Gray, and Matt Berlin.
\newblock An embodied cognition approach to mindreading skills for socially
  intelligent robots.
\newblock {\em The Internat'l Journal of Robotics Research}, 28(5):656--680,
  2009.

\bibitem{10.1109/cvpr52688.2022.02026}
De'Aira Bryant, Siqi Deng, Nashlie Sephus, Wei Xia, and Pietro Perona.
\newblock {Multi-Dimensional, Nuanced and Subjective – Measuring the
  Perception of Facial Expressions}.
\newblock {\em 2022 IEEE/CVF Conference on CVPR}, 00:20900--20909, 2022.

\bibitem{Butepage2017CVPR}
Judith Butepage, Michael~J. Black, Danica Kragic, and Hedvig Kjellstrom.
\newblock Deep representation learning for human motion prediction and
  classification.
\newblock In {\em Proceedings of the IEEE Conference on Computer Vision and
  Pattern Recognition (CVPR)}, July 2017.

\bibitem{8765346}
Z.~{Cao}, G.~{Hidalgo Martinez}, T.~{Simon}, S.~{Wei}, and Y.~A. {Sheikh}.
\newblock Openpose: Realtime multi-person 2d pose estimation using part
  affinity fields.
\newblock {\em IEEE Trans. on Pattern Analysis and Machine Intell.}, 2019.

\bibitem{cao2017realtime}
Zhe Cao, Tomas Simon, Shih-En Wei, and Yaser Sheikh.
\newblock Realtime multi-person 2d pose estimation using part affinity fields.
\newblock In {\em Proc. of the IEEE conference on CVPR}, pages 7291--7299,
  2017.

\bibitem{dautenhahn2007socially}
Kerstin Dautenhahn.
\newblock Socially intelligent robots: dimensions of human--robot interaction.
\newblock {\em Philosophical transactions of the royal society B: Biological
  sciences}, 362(1480):679--704, 2007.

\bibitem{dosovitskiy2020image}
Alexey Dosovitskiy, Lucas Beyer, Alexander Kolesnikov, Dirk Weissenborn,
  Xiaohua Zhai, Thomas Unterthiner, Mostafa Dehghani, Matthias Minderer, Georg
  Heigold, Sylvain Gelly, et~al.
\newblock An image is worth 16x16 words: Transformers for image recognition at
  scale.
\newblock In {\em Internat'l Conference on Learning Representations}, 2020.

\bibitem{ekman1972hand}
Paul Ekman and Wallace~V Friesen.
\newblock Hand movements.
\newblock {\em Journal of communication}, 22(4):353--374, 1972.

\bibitem{fang2022alphapose}
Hao-Shu Fang, Jiefeng Li, Hongyang Tang, Chao Xu, Haoyi Zhu, Yuliang Xiu,
  Yong-Lu Li, and Cewu Lu.
\newblock Alphapose: Whole-body regional multi-person pose estimation and
  tracking in real-time.
\newblock {\em arXiv preprint arXiv:2211.03375}, 2022.

\bibitem{10.1109/lra.2019.2930434}
Panagiotis~Paraskevas Filntisis, Niki Efthymiou, Petros Koutras, Gerasimos
  Potamianos, and Petros Maragos.
\newblock {Fusing Body Posture With Facial Expressions for Joint Recognition of
  Affect in ChildRobot Interaction}.
\newblock {\em IEEE Robotics and Autom. Letters}, 4(4):4011--4018, 2019.

\bibitem{resnet}
Kaiming He, Xiangyu Zhang, Shaoqing Ren, and Jian Sun.
\newblock Deep residual learning for image recognition.
\newblock In {\em Proceedings of the IEEE conference on CVPR}, pages 770--778,
  2016.

\bibitem{huang2021emotion}
Yibo Huang, Hongqian Wen, Linbo Qing, Rulong Jin, and Leiming Xiao.
\newblock Emotion recognition based on body and context fusion in the wild.
\newblock In {\em Proceedings of the IEEE/CVF Internat'l Conference on Computer
  Vision}, pages 3609--3617, 2021.

\bibitem{10.1109/iccvw54120.2021.00403}
Yibo Huang, Hongqian Wen, Linbo Qing, Rulong Jin, and Leiming Xiao.
\newblock {Emotion Recognition Based on Body and Context Fusion in the Wild}.
\newblock {\em 2021 IEEE/CVF Internat'l Conference on Computer Vision Workshops
  (ICCVW)}, 00:3602--3610, 2021.

\bibitem{10.5220/0010359506690679}
Chaudhary Ilyas, Rita Nunes, Kamal Nasrollahi, Matthias Rehm, and Thomas
  Moeslund.
\newblock {Deep Emotion Recognition through Upper Body Movements and Facial
  Expression}.
\newblock {\em Proceedings of the 16th Internat'l Joint Conference on Computer
  Vision, Imaging and Computer Graphics Theory and Applications}, pages
  669--679, 2021.

\bibitem{Kanazawa2019CVPR}
Angjoo Kanazawa, Jason~Y. Zhang, Panna Felsen, and Jitendra Malik.
\newblock Learning 3d human dynamics from video.
\newblock In {\em Proceedings of the IEEE/CVF Conference on CVPR}, June 2019.

\bibitem{kollias2022abaw}
Dimitrios Kollias.
\newblock Abaw: Valence-arousal estimation, expression recognition, action unit
  detection \& multi-task learning challenges.
\newblock In {\em Proceedings of the IEEE/CVF Conference on Computer Vision and
  Pattern Recognition}, pages 2328--2336, 2022.

\bibitem{kosti2017emotion}
Ronak Kosti, Jose~M Alvarez, Adria Recasens, and Agata Lapedriza.
\newblock Emotion recognition in context.
\newblock In {\em Proceedings of the IEEE conference on computer vision and
  pattern recognition}, pages 1667--1675, 2017.

\bibitem{kosti2019context}
Ronak Kosti, Jose~M Alvarez, Adria Recasens, and Agata Lapedriza.
\newblock Context based emotion recognition using emotic dataset.
\newblock {\em IEEE Trans. on pattern anal. and machine intelligence},
  42(11):2755--2766, 2019.

\bibitem{li2021hybrik}
Jiefeng Li, Chao Xu, Zhicun Chen, Siyuan Bian, Lixin Yang, and Cewu Lu.
\newblock Hybrik: A hybrid analytical-neural inverse kinematics solution for 3d
  human pose and shape estimation.
\newblock In {\em Proceedings of the IEEE/CVF Conference on Computer Vision and
  Pattern Recognition}, pages 3383--3393, 2021.

\bibitem{liu2021swin}
Ze~Liu, Yutong Lin, Yue Cao, Han Hu, Yixuan Wei, Zheng Zhang, Stephen Lin, and
  Baining Guo.
\newblock Swin transformer: Hierarchical vision transformer using shifted
  windows.
\newblock In {\em Proceedings of the IEEE/CVF Internat'l Conf. on Computer
  Vision}, pages 10012--10022, 2021.

\bibitem{luo2021towards}
Yu~Luo.
\newblock {\em Towards Automated Recognition of Bodily Expression of Emotion in
  the Wild}.
\newblock The Pennsylvania State University, 2021.

\bibitem{mazhar2018}
Osama Mazhar, Sofiane Ramdani, Benjamin Navarro, Robin Passama, and Andrea
  Cherubini.
\newblock Towards real-time physical human-robot interaction using skeleton
  information and hand gestures.
\newblock In {\em 2018 IEEE/RSJ Internat'l Conference on Intelligent Robots and
  Systems (IROS)}, pages 1--6, 2018.

\bibitem{miller2019causal}
Lynn~C Miller, Sonia~Jawaid Shaikh, David~C Jeong, Liyuan Wang, Traci~K Gillig,
  Carlos~G Godoy, Paul~R Appleby, Charisse~L Corsbie-Massay, Stacy Marsella,
  John~L Christensen, and Stephen~J Read.
\newblock Causal inference in generalizable environments: Systematic
  representative design.
\newblock {\em Psychological Inquiry}, 30(4):173--202, 2019.

\bibitem{mumm2011human}
Jonathan Mumm and Bilge Mutlu.
\newblock Human-robot proxemics: physical and psychological distancing in
  human-robot interaction.
\newblock In {\em Proceedings of the 6th Internat'l conference on Human-robot
  interaction}, pages 331--338. ACM, 2011.

\bibitem{nakata1998expression}
Toru Nakata, Tomomasa Sato, Taketoshi Mori, and Hiroshi Mizoguchi.
\newblock Expression of emotion and intention by robot body movement.
\newblock In {\em Proceedings of the 5th Internat'l Conf. on autonomous
  systems}, 1998.

\bibitem{10.1109/iciev.2019.8858536}
Yuki Ono, Saizo Aoyagi, Yoichi Yamazaki, Michiya Yamamoto, and Noriko Nagata.
\newblock {Emotion Estimation Using Body Expression Types Based on LMA and
  Sensitivity Analysis}.
\newblock {\em 2019 Joint 8th Internat'l Conference on Informatics, Electronics
  \& Vision (ICIEV) and 2019 3rd Internat'l Conference on Imaging, Vision \&
  Pattern Recognition (icIVPR)}, 00:348--353, 2019.

\bibitem{park2012law}
Eunil Park, Dallae Jin, and Angel~P del Pobil.
\newblock The law of attraction in human-robot interaction.
\newblock {\em Internat'l Journal of Advanced Robotic Systems}, 9(2):35, 2012.

\bibitem{pishchulin2012articulated}
Leonid Pishchulin, Arjun Jain, Mykhaylo Andriluka, Thorsten Thorm{\"a}hlen, and
  Bernt Schiele.
\newblock Articulated people detection and pose estimation: Reshaping the
  future.
\newblock In {\em 2012 IEEE Conference on CVPR}, pages 3178--3185. IEEE, 2012.

\bibitem{radford2021learning}
Alec Radford, Jong~Wook Kim, Chris Hallacy, Aditya Ramesh, Gabriel Goh,
  Sandhini Agarwal, Girish Sastry, Amanda Askell, Pamela Mishkin, Jack Clark,
  et~al.
\newblock Learning transferable visual models from natural language
  supervision.
\newblock In {\em Internat'l Conference on Machine Learning}, pages 8748--8763.
  PMLR, 2021.

\bibitem{rauthmann2014situational}
John~F Rauthmann, David Gallardo-Pujol, Esther~M Guillaume, Elysia Todd,
  Christopher~S Nave, Ryne~A Sherman, Matthias Ziegler, Ashley~Bell Jones, and
  David~C Funder.
\newblock The situational eight diamonds: A taxonomy of major dimensions of
  situation characteristics.
\newblock {\em Journal of Personality and Social Psychology}, 107(4):677, 2014.

\bibitem{redmon2016you}
Joseph Redmon, Santosh Divvala, Ross Girshick, and Ali Farhadi.
\newblock You only look once: Unified, real-time object detection.
\newblock In {\em Proceedings of the IEEE conference on computer vision and
  pattern recognition}, pages 779--788, 2016.

\bibitem{ruffaldi2016third}
Emanuele Ruffaldi, Filippo Brizzi, Franco Tecchia, and Sandro Bacinelli.
\newblock Third point of view augmented reality for robot intentions
  visualization.
\newblock In {\em Internat'l Conference on Augmented Reality, Virtual Reality
  and Computer Graphics}, pages 471--478. Springer, 2016.

\bibitem{PSTMO}
Wenkang Shan, Zhenhua Liu, Xinfeng Zhang, Shanshe Wang, Siwei Ma, and Wen Gao.
\newblock {P-STMO: Pre-Trained Spatial Temporal Many-to-One Model for 3D Human
  Pose Estimation}.
\newblock {\em arXiv}, 2022.

\bibitem{shan2022p}
Wenkang Shan, Zhenhua Liu, Xinfeng Zhang, Shanshe Wang, Siwei Ma, and Wen Gao.
\newblock P-stmo: Pre-trained spatial temporal many-to-one model for 3d human
  pose estimation.
\newblock In {\em European Conference on Computer Vision}, pages 461--478.
  Springer, 2022.

\bibitem{stoeva2020}
Darja Stoeva and Margrit Gelautz.
\newblock Body language in affective human-robot interaction.
\newblock In {\em Companion of the 2020 ACM/IEEE Internat'l Conference on
  Human-Robot Interaction}, HRI '20, page 606–608, New York, NY, USA, 2020.
  Association for Computing Machinery.

\bibitem{takayama2009influences}
Leila Takayama and Caroline Pantofaru.
\newblock Influences on proxemic behaviors in human-robot interaction.
\newblock In {\em Intelligent robots and systems, 2009. IROS 2009. IEEE/RSJ
  Internat'l conference on}, pages 5495--5502. IEEE, 2009.

\bibitem{Tang_2023_CVPR}
Zhenhua Tang, Zhaofan Qiu, Yanbin Hao, Richang Hong, and Ting Yao.
\newblock 3d human pose estimation with spatio-temporal criss-cross attention.
\newblock In {\em Proceedings of the IEEE/CVF Conference on Computer Vision and
  Pattern Recognition (CVPR)}, pages 4790--4799, June 2023.

\bibitem{van2013biological}
Jeroen~JA van Boxtel and Hongjing Lu.
\newblock A biological motion toolbox for reading, displaying, and manipulating
  motion capture data in research settings.
\newblock {\em Journal of vision}, 13(12):7--7, 2013.

\bibitem{attention2017}
Ashish Vaswani, Noam Shazeer, Niki Parmar, Jakob Uszkoreit, Llion Jones,
  Aidan~N Gomez, {\L}ukasz Kaiser, and Illia Polosukhin.
\newblock Attention is all you need.
\newblock {\em Advances in neural information processing systems}, 30, 2017.

\bibitem{tuyen2018}
Nguyen~Tan Viet~Tuyen, Sungmoon Jeong, and Nak~Young Chong.
\newblock Emotional bodily expressions for culturally competent robots through
  long term human-robot interaction.
\newblock In {\em 2018 IEEE/RSJ Internat'l Conf. on Intelligent Robots and
  Systems}, pages 2008--2013, 2018.

\bibitem{volkova2014mpi}
Ekaterina Volkova, Stephan De~La~Rosa, Heinrich~H B{\"u}lthoff, and Betty
  Mohler.
\newblock The mpi emotional body expressions database for narrative scenarios.
\newblock {\em PloS one}, 9(12):e113647, 2014.

\bibitem{xu_mdan_2022}
Liwen Xu, Zhengtao Wang, Bin Wu, and Simon Lui.
\newblock Mdan: Multi-level dependent attention network for visual emotion
  analysis.
\newblock In {\em Proceedings of the IEEE/CVF Conference on Computer Vision and
  Pattern Recognition}, pages 9479--9488, 2022.

\bibitem{yang2022seeking}
Jingyuan Yang, Jie Li, Leida Li, Xiumei Wang, Yuxuan Ding, and Xinbo Gao.
\newblock Seeking subjectivity in visual emotion distribution learning.
\newblock {\em IEEE Transactions on Image Processing}, 31:5189--5202, 2022.

\bibitem{Yang_2018_CVPR}
Jufeng Yang, Dongyu She, Yu-Kun Lai, Paul~L. Rosin, and Ming-Hsuan Yang.
\newblock Weakly supervised coupled networks for visual sentiment analysis.
\newblock In {\em The IEEE Conference on Computer Vision and Pattern
  Recognition}, 2018.

\bibitem{zhao2019pdanet}
Sicheng Zhao, Zizhou Jia, Hui Chen, Leida Li, Guiguang Ding, and Kurt Keutzer.
\newblock Pdanet: Polarity-consistent deep attention network for fine-grained
  visual emotion regression.
\newblock In {\em Proceedings of the 27th ACM Internat'l conference on
  multimedia}, pages 192--201, 2019.

\bibitem{zhou2018auto}
Yi~Zhou, Zimo Li, Shuangjiu Xiao, Chong He, Zeng Huang, and Hao Li.
\newblock Auto-conditioned recurrent networks for extended complex human motion
  synthesis.
\newblock In {\em Internat'l Conference on Learning Representations}, 2018.

\end{thebibliography}

\end{document}